# Tracking daily paths in home contexts with RSSI fingerprinting based on UWB through deep learning models

A. Polo-Rodríguez[1] · J. C. Valera[2] · J. Peral[3] · D. Gil[2]  · J. Medina-Quero[1]

## Abstract

The field of human activity recognition has evolved significantly, driven largely by advancements in Internet of Things (IoT) device technology, particularly in personal devices. This study investigates the use of ultra-wideband (UWB) technology for tracking inhabitant paths in home environments using deep learning models. UWB technology estimates user locations via time-of-flight and time-difference-of-arrival methods, which are significantly affected by the presence of walls and obstacles in real environments, reducing their precision. To address these challenges, we propose a fingerprinting-based approach utilizing received signal strength indicator (RSSI) data collected from inhabitants in two flats (60 m$^2$ and 100 m$^2$) while performing daily activities. We compare the performance of convolutional neural network (CNN), long short-term memory (LSTM), and hybrid CNN+LSTM models, as well as the use of Bluetooth technology. Additionally, we evaluate the impact of the type and duration of the temporal window (future, past, or a combination of both). Our results demonstrate a mean absolute error close to 50 cm, highlighting the superiority of the hybrid model in providing accurate location estimates, thus facilitating its application in daily human activity recognition in residential settings.

✉ J. Peral
jperal@dlsi.ua.es

✉ D. Gil
dgil@dtic.ua.es

A. Polo-Rodríguez
auro@ugr.es

J. C. Valera
valeramanzanojc@gmail.com

J. Medina-Quero
javiermq@ugr.es

[1] Department of Computer Engineering, Automation and Robotics, University of Granada, 18071 Granada, Spain

[2] Department of Computer Technology and Computation, University of Alicante (UA), 03690 Alicante, Spain

[3] Department of Software and Computing Systems, University of Alicante, 03690 Alicante, Spain

## 1 Introduction

Significant technological advances have recently enabled successful research in human activity recognition (HAR), offering solutions to challenges in ubiquitous computing, human-computer interaction, and human behaviour analysis [1]. This research area has benefited from the advent of wearable sensors that collect data to aid monitoring. Devices like smart wearables or mobile phones [2] are equipped with sensors (accelerometer, gyroscope, magnetometer, Bluetooth, Wi-Fi, microphones, proximity and light sensors, heart rate monitors or GPS) that capture various specific types of information. In addition to wearable sensors, some sensors collect environmental data through devices fixed in specific locations. These are known as ambient sensors, examples of which include cameras or acoustic and thermal sensors. Both wearables and ambient sensors ubiquitously collect raw data from their targets. Analysing these data will generate knowledge to solve complex problems [3].

A critical challenge in the domain of HAR is the identification of the specific individual performing an activity in scenarios involving multiple individuals; this complexity is exemplified in environments such as multi-resident apartments [4]. In such contexts, indoor location tracking methodologies are utilized to accurately identify and monitor the trajectories of entities, encompassing both individuals and objects, within indoor settings [5]. Indoor positioning has become an essential technology in various applications, including smart homes, healthcare, and assisted living. Multiple different technologies can be used for indoor tracking and positioning [6]: (1) radio-based (ultra-wideband –UWB–, Wi-Fi, Bluetooth, etc.); (2) optical (video camera, infrared, etc.); (3) magnetic (magnetic strength); or (4) acoustic (ultrasound)[1]. Despite its importance, several technical gaps persist in the field. One major issue is the lack of extensive datasets collected in real-world environments. Most experimental setups and evaluations are conducted in controlled laboratory settings, which do not accurately represent the complexities and variabilities of actual indoor environments [7]. Historically, indoor positioning systems have primarily relied on Bluetooth low energy (BLE) technology due to its low cost and ease of deployment. BLE is widely integrated into wearables and smartphones, which can act as tags, ambient beacons, or anchors collecting received signal strength indicators (RSSI) values [8]. However, the variability of BLE signals and the dynamic nature of indoor spaces, influenced by furniture and other obstacles, require a high density of beacons to achieve accurate positioning [9, 10].

Fingerprinting, a common technique used in location systems, involves creating a map of signal strengths at various locations and using this map to estimate positions [11, 12]. Although effective, this method requires a substantial amount of labelled data, and the performance of BLE can be inconsistent due to signal fluctuations. In recent years, UWB technology has emerged as a promising alternative for indoor positioning, offering higher precision than BLE, as it allows us to locate with coordinate precision rather than area precision. However, UWB technology also faces significant challenges that affect its performance. Signal obstruction is a primary concern, as UWB signals are susceptible to attenuation and multipath effects caused by physical obstructions such as walls and furniture. This leads to reduced accuracy in location estimation [13]. In addition, effective UWB deployment requires a significant number of strategically placed anchors to ensure complete coverage and accuracy. Each anchor must maintain a clear line of sight to other anchors to provide precise location coordinates, which can be logistically challenging and costly [14]. To address the problem of nonline-of-sight (NLOS) conditions and reduce the impact of the number of anchors required, the fingerprinting method plays a crucial role [15]. By correlating the real-world coordinates

---

[1] https://www.sewio.net/indoor-location-tracking-and-positioning/ (visited on 26 November 2022)

with the RSSI values received from each beacon, fingerprinting can provide accurate location estimates even in complex environments. This method mitigates the limitations of UWB in obstructed settings by leveraging the consistent patterns of signal strengths mapped during the fingerprinting process.

This work is motivated by the need to develop robust, accurate, and cost-effective indoor positioning systems to track entities (persons or objects) within indoor environments. Traditional methods often struggle with signal obstruction, environmental variability, and high infrastructure costs. To address these challenges and improve precision, we focus on radio-based technologies, specifically UWB and BLE. Our approach involves deploying anchors throughout the facility to receive signals from tags attached to the entities. Initially, the user maps their location within the environment, and these labels are collected and correlated with the signals received by the anchors. These data are sent to a server for processing, where the model learns from the labelled data. Once trained, the system can accurately calculate the location in real time of the entity using only RSSI signals, reducing costs and facilitating deployment in real environments. In Fig. 1 we detail the flow of the work. In summary, these key points are addressed:

- We have evaluated UWB deployments, including a comparative analysis with Bluetooth for inhabitant location. By comparing UWB with BLE, we aim to determine the most suitable technology for different scenarios. While UWB offers high precision with coordinate-based localization, BLE can be sufficient for applications requiring area-based localization. This comparison helps in selecting the appropriate technology based on the specific needs of the environment.
- Fingerprinting techniques have been integrated instead of trilateration or triangulation to deduce the cost of anchor deployment. Learning from the RSSI measurements in an instant of time is proposed in conjunction with others previously saved by the sliding window method.
- Deep Learning models with convolutional neural network (CNN) and long short-term memory (LSTM) for regression of the user's location have been evaluated.
- An ad hoc case study in two flats and different configurations has been deployed, including the ground truth labelled in real time by the inhabitant using a tablet application. The case studies have been designed to be deployed in a real environment with dynamic tracks representing the daily human activity of each inhabitant under natural conditions.
- A comprehensive evaluation and comparison of several machine and deep learning techniques (CNN, LSTM, LSTM+CNN, RF –random forest–, and SVM –support vector machine–) was conducted to identify the most accurate method for predicting the user's location. Additionally, the influence of window type (past, future, and past+future) and window size was assessed.

The remainder of this paper is structured as follows. Section 2 reviews the related work. Section 3 presents our approach, including the type of devices, the data processing, and the deep learning model. Section 4 summarises a case study that includes experimentation with real scenarios. Finally, in Section 5, the conclusions and future work are presented.

## 2 Background

This section details the background and work related to the proposal presented here. Depending on the type of sensing method for modelling HAR, we highlight the following three groups:

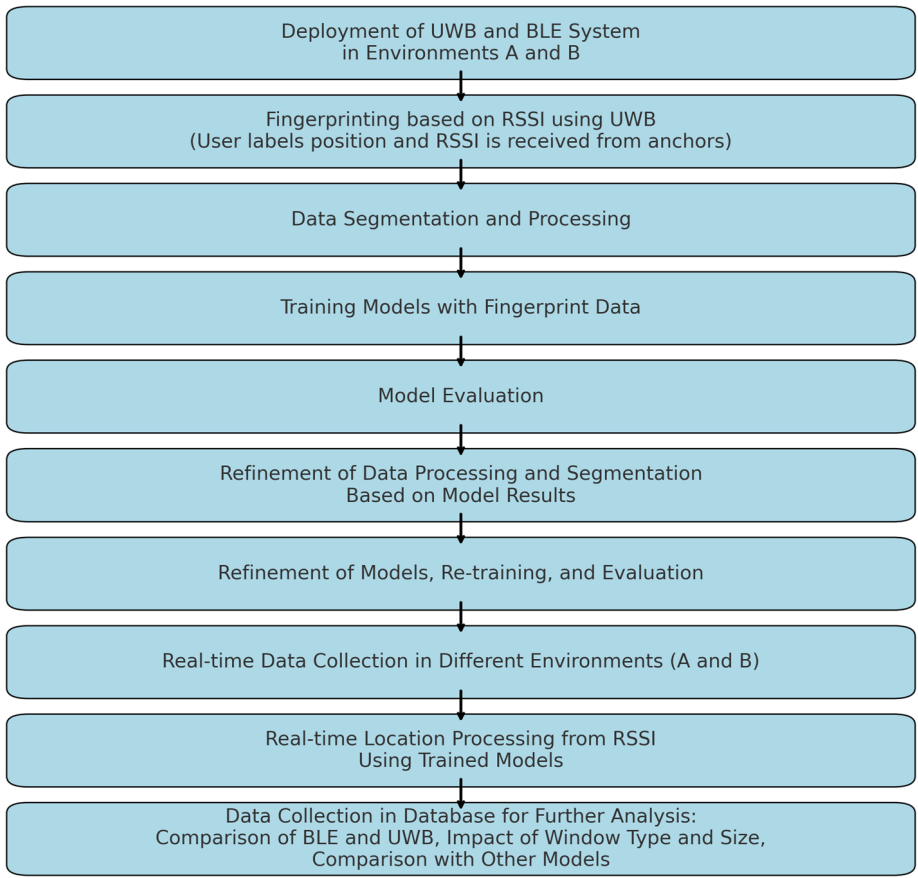

**Fig. 1** Work flow of our indoor positioning proposal

1. Vision-based HAR, where vision sensors are commonly fitted inside a smart home or a building. These have the advantage of offering essential information about the environment and movement. Among their disadvantages, they can suffer considerable variations in accuracy due to changes in light, and care must be taken not to violate people's privacy when deployed [16, 17].
2. Ambient sensor-based HAR, such as presence or open/close sensors (binary sensors), acoustic or multimodal sensors. The main advantage of these is that they are less intrusive than the rest of the sensors, while a disadvantage is the difficulty of deployment (only possible in smart homes) [18].
3. Wearable sensor-based HAR is the most popular choice thanks to the emergence of smartphones and smartwatches that include accelerometers, gyroscopes, microphones, Bluetooth, Wi-Fi, etc. The main drawback is their comparatively short battery life and the possibility that the study subjects forget to put the device on [19].

After the data is collected from the sensors, you can apply several HAR methods such as machine learning (ML), template matching (TM), or event-based (EB) methods. Among the literature where EB HAR methods are used, we can find articles such as [20, 21], and [22], which focus on recognising simple activities such as walking, waving, or clapping. In the group of studies using TM, we have [23] and [24], which focus on the recognition of simple activities and ADL (activities of daily living, such as dressing, bathing, etc.), or [25], which analyses falls in addition to simple activities. In terms of the number of articles, machine learning is the most highly researched area of the three. Among these, we find [26–29], and [30] that focus on the recognition of ADL. Despite this, it is still an underexplored field in which it is difficult to improve accuracy under more realistic conditions [31]. Other major challenges include choosing the set of attributes to be measured, collecting datasets under real-world conditions, and creating a portable and affordable data collection system that supports multi-occupancy [32].

In this regard, indoor location tracking is vital since current sensors might have false or incorrect activations [33]. The scientific community has addressed this issue through different approaches. Some studies have used passive infrared (PIR) sensors to detect the position of an inhabitant inside a smart home, as in [34], although user identification is limited. It is also common to use thermal vision sensors, as they are less invasive than visible spectrum ones, as the authors in [35], who used low-resolution thermal cameras. However, in multi-occupancy environments, detecting which person is performing the activity is critical. To do this, it is necessary to know the location of each user [36, 37]. There are several techniques to estimate the position of the inhabitant, and one of the most widespread is the use of RSSI. It is a measure of the power present in a received radio signal. It does not require pairing, is low-cost and offers good performance [38]. Nevertheless, RSSI values are influenced by several factors, including the level of battery of the device, which can affect signal strength, and hardware variations between different devices. Environmental factors such as obstacles, interference from other electronic devices, and multipath effects also contribute to RSSI variability However, sensor fusion is useful for more reliable user identification [39]. As delineated in the preceding section, the predominant wireless technology employed for indoor localisation is BLE. For instance, [40] create an ecosystem consisting of Beacon devices, Bluetooth smart devices, and Wi-Fi access points to develop an efficient method of indoor location determination using Wi-Fi and BLE technologies in a hybrid way. The authors of [41] give a comprehensive survey of numerous indoor positioning systems, with summaries and comparison in security and privacy, cost, performance and robustness.

Fingerprinting is also a prevalent technique for tracking individuals, as evidenced in [42–44]. This method can be integrated with RSSI or other deep learning methodologies to develop models for location estimation. In addition to BLE scanners and beacons, [45] uses fingerprinting by amalgamating the results of RSSI triangulation with those derived from fingerprinting. This methodology is adopted in [46] and [12], with the former introducing fingerprint positioning technologies and the latter employing machine learning and intelligent algorithms for fingerprinting. The primary challenge with BLE is the variability of the signal [9] and the variability of indoor environments due to furniture or decorative objects, which requires the installation of a high density of beacons [10]. Some authors of this work have previously conducted studies on this topic [47] and possess the required experience to execute this proposal successfully. Regarding time range approaches, although ToA (time of arrival) offers superior performance [48], it involves higher computational processing, synchronisation, and pairing requirements. Furthermore, substantial communication overhead and the need for appropriate synchronisation circuits between devices raise the cost and maintenance [49]. To estimate the position of the inhabitant from these signals, vari-ous proposals have employed techniques such as trilateration and fingerprinting [50–52]. A combination of both techniques is frequently observed, as trilateration has limitations when the number of beacons is low or they do not communicate effectively due to environmental noise (walls or obstacles). However, fingerprinting requires a substantial amount of labelled data per environment [53]. UWB technology provides high precision for indoor positioning but encounters several challenges that affect its performance, such as signal obstruction, the requirement for a large number of strategically placed anchors to ensure coverage and accuracy, and the high cost of the UWB infrastructure [54]. Consequently, in this work, we have opted for location estimation based on RSSI signals for the following reasons: i) the hardware is simple and cost-effective as it requires only a minimal number of anchors [55]; ii) the lower power requirement since it is available in the physical layer [56]; iii) time-based ranging necessitates additional hardware, such as a directional antenna or external signal sources, which increases energy consumption and hardware resources [57]; iv) time-based methods are designed for line-of-sight conditions and empty spaces, whereas RSSI yields better results in dynamic paths and obstructed environments [58].

To enhance the accuracy of indoor positioning systems, incorporating advanced deep learning architectures, such as CNNs and LSTMs, with fingerprinting techniques is highly beneficial. CNNs are proficient in detecting spatial patterns in RSSI data, while LSTMs are adept at capturing temporal dependencies and dynamics in signal sequences. Using these deep learning models improves location estimation accuracy, better adapts to the variability of indoor environments, and reduces the reliance on a dense network of anchors. This combination allows for the creation of more sophisticated and adaptive models that can provide precise real-time positioning in complex indoor settings. This can be corroborated in a variety of works in different areas. The study on oil well production prediction based on a CNN-LSTM model highlights the potential of combining spatial and temporal data, which can be applied directly to our domain [59]. The hybrid wavelet group method for fault prediction in electrical power insulators demonstrates the utility of handling data intricacies, similar to the complex nature of indoor signal data. To compare and validate this method, a benchmark with LSTM and the Adaptive Neuro-Fuzzy Inference System (ANFIS) was carried out [60]. The work by [61] underscores the importance of attention mechanisms combined with CNN and LSTM to improve prediction accuracy, which could benefit the refinement of our models. Similarly, [62] integrates the attention mechanism with an LSTM recurrent network to enhance prediction. Finally, [63] illustrates a comprehensive approach to combining multiple models, which can inspire more robust framework development for our application.

Regarding the state-of-the-art of the models using UWB technology, Table 1 summarises the most relevant works of the recent years due to the rapid evolution of this technology, which is continuously undergoing significant changes and improvements. This comparative analysis highlights the advancements and varied applications of UWB technology in different environments, showcasing the key findings, strengths and limitations of each study.

In summary, our study includes the use of UWB technology for position estimation through RSSI signals, along with a fingerprinting technique to improve accuracy. This system is particularly promising, because of its low computational demands and respect for the privacy of the inhabitants. Previous research has validated the effectiveness of this technology, demonstrating superior accuracy and performance [69]. Our proposal advances this state-of-the-art by significantly reducing computational costs and ensuring privacy. Additionally, a unique feature of our study is the use of a manually labelled dataset, collected in a real world setting where everyday human activities occur naturally.

Table 1 Summary of the UWB state-of-the-art approaches

| | Key findings | Strengths | Limitations |
|---|---|---|---|
| Alarifi et al. (2016) [64] | UWB systems provide high accuracy for indoor positioning, outperforming many other wireless technologies in complex environments. | Comprehensive review of various UWB techniques, detailed analysis of algorithms and implementations, and comparison with other technologies. | Mostly theoretical with limited experimental validation; emphasises a broad analysis but lacks specific performance benchmarks; focuses on existing methods without proposing new solutions. |
| Jiménez Ruiz et al. (2017) [65] | The study provides a thorough comparison of three leading UWB systems, offering valuable insights into their strengths and weaknesses. | Comprehensive comparison, real-world testing, and diverse performance metrics. | Environment specificity (results are based on specific indoor environments), potential outdated technology, and focus on commercial systems. |
| Großwindhager et al. (2019) [66] | SnapLoc: provides ultra-fast localisation, can handle an unlimited number of tags, and demonstrates robust performance in various indoor environments. | Speed, scalability, high accuracy, and versatility in different settings. | Complex implementation, specific use cases, and potential impact of environmental factors. |
| Wang et al. (2023) [67] | The integration of UWB with IMU (inertial measurement unit) devices improves positioning accuracy and reliability in challenging environments like underground mines where GPS cannot be used. | Innovative combination of UWB and IMU devices, detailed experimental validation in real-world scenarios, and robust performance in harsh conditions. | High cost and complexity of base station deployment, and simulation experiments in a limited area. |
| Qian et al. (2024) [68] | The UWB tracker demonstrated high accuracy in in-home positioning, proving to be a reliable tool for monitoring the movements of older adults within their living environments. | High positioning accuracy and potential for improving the safety and autonomy of older adults by enabling continuous monitoring. | The study may be limited by the specific home environments tested, and further research is needed to validate the tracker's performance in different types of environments. |

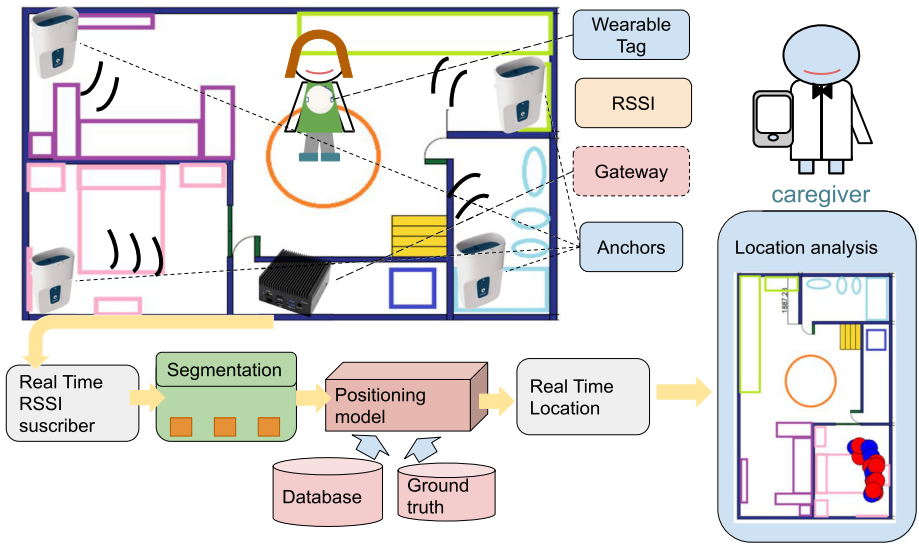

**Fig. 2** Components integrated into the proposed architecture to compute location in real time from RSSI tag data to UWB anchors using the fingerprinting model

## 3 Materials and methods

This section describes the architecture and approach used to estimate user indoor location, detailing the type of devices used and the data processing method, as well as providing a description of the deep learning model. In this context, the architecture is composed of a smart environment with anchors located on the walls and ceiling, and the user wears a tag that provides RSSI data to the anchors. The RSSI values collected by the anchors are published in real time to a data subscriber, which segments and pre-processes the data under a sliding window approach to fit the input to the location model. The location model provides the 2D location for each inhabitant as output. Models the prediction based on the ground truth and RSSI data from the data labelled with the knowledge base. In the end, the location of the residents is stored in a database to provide user habitation areas to caregivers remotely and in real time. In Fig. 2, we describe the components proposed to compute the location of inhabitants in real time from RSSI tag data to UWB anchors.

In the following sections, we detail the materials and methods used for segmentation and the learning models that make up our proposal.

### 3.1 RTLS based on UWB technology

This section describes a sensor kit that comprises a real-time location system (RTLS). This technology has recently been used in the scientific community for indoor position tracking [70]. This has led to their integration into smartphones and other devices with high location accuracy [71], especially Samsung, Google, Apple, and Xiaomi with Galaxy, Pixel, iPhone, Apple Watch, and MIX4, respectively. Some security issues remain open due to the ability of UWB to accurately track people indoors. The devices used in this work are part of the Enterprise Kit Lite[2], which uses trilateration to process up to 100 positions per second.

---

[2] https://www.pozyx.io (visited on 26 November 2022).

This system cannot always compute the position; it depends on the number and location of the devices. The base price is around €3,000, including a gateway, four anchors, and four wearable tags. Figure 3 shows the device.

- **Anchors.** The anchors are the indispensable element for successful positioning. They capture the signals, process the data, and send them to the platform. The system requires at least three anchors visible to each other to calculate 2D positions. It supports two protocols (TDOA –time difference of arrival– and TWR –two-way-ranging–) for positioning. The anchors are available in IP20 or IP66/67 versions and are adapted to VESA mounts for walls and ceilings. Anchors can be powered with a DC jack via power over Ethernet (PoE, IEEE 802.3af) or PoE+ (IEEE 802.3at) with a maximum of 4 anchors in a daisy chain.
- **Wearable tags.** They are small devices measuring 66mm x 65.4mm x 17mm (L × W × H) with a battery life of up to 5 years and the possibility to replace the battery. They have ultra-low power consumption and a maximum refresh rate of 10 Hz. The tags have NFC connectivity for activation, installation and configuration, and they can modify their ID to make them easier to identify. They are compatible with the TDOA positioning system described above, and they have a 3-axis accelerometer as an additional sensor and an IP66/67 rating.

### 3.2 Segmentation of RSSI data

Following a formal definition, an anchor $s$ detects the presence of a tag $e$ in real time in the form of a pair $\overline{(s,e)_i} = \{s_i, t_i\}$, where $(s,e)_i$ represents RSSI and $t_i$ the timestamp. Thus, we obtain a data stream for a given anchor and tag $(s,e)$, which is defined by $\overline{S_{(s,e)}} = \{\overline{(s,e)_0}, \ldots, \overline{(s,e)_i}\}$ and a given value in a timestamp $t_i$ by $S_{(s,e)}(t_i) = s_i$.

We define several symmetric temporal sliding windows to homogenise the data collected by the different sensors. They are defined by the window size of a time interval $W_w = [W_w^-, W_w^+]$, segment the samples of a RSSI sensor stream $\overline{S_{(s,e)}}$ and aggregate the values

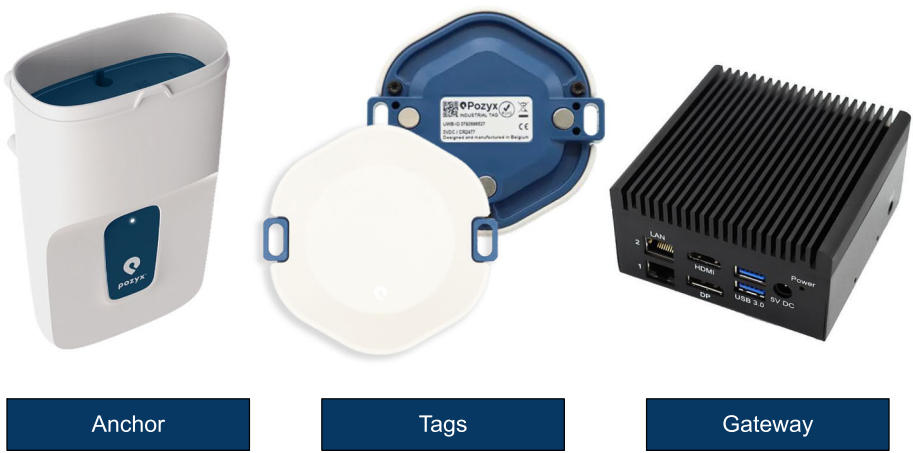

**Fig. 3** Pozyx system components: anchors, tags and gateway

$\overline{(s,e)_i}$ using several aggregation functions $T_t(S_{(s,e)}, W_w, t^*)$:

$$T_t(S_{(s,e)}, W_w, t^*) = \bigcup_{\overline{(s,e)_i}}^{S(s,e)} s_i, t_i \in [t^* - W_w^-, t^* - W_w^+] \quad (1)$$

Therefore, aggregation functions $\cup$ are applied to the data $(s, e)_i$ within a time interval $W_w = [W_w^-, W_w^+]$ at a given point in time $t^*$. Specifically, the following aggregation functions are proposed in this work $\cup = \{\mu, \max, \min\}$, which are proposed as key descriptors for high-rate streams [72]. These aggregation functions provide a strong representation of segmented data by time intervals to homogenise the heterogeneous raw sensor data collected at different collection rates.

Using data aggregations, signal segmentation is defined by several short sliding temporal windows by consecutive time intervals where $W = \{W_1 = [W_1^-, W_1^+], \ldots, W_i = [W_i^-, W_i^+]\}$, $W_i^- = W_{i-1}^+$.

In this way, we obtain a sequence of features from sensor data streams for each tag $e$, whose shape is $|S| \times |\cup| \times |W|$:

$$S_e^*(t^*) \rightarrow \begin{cases} S_e^1(t^*) = T_t(S_{(S1,e)}, [W_1^-, W_1^+], t^*)\ldots, \\ \rightarrow T_t(S_{(S1,e)}, [W_i^-, W_i^+], t^*) \\ \ldots \\ S_e^s(t^*) = T_t(S_{(s,e)}, [W_1^-, W_1^+)\ldots, \\ \rightarrow T_t(S_{(s,e)}, [W_i^-, W_i^+], t^*) \end{cases}$$

At this point, the sensor streams $S_e^s(t^*)$ for each tag $e$ and anchor $s$ are: i) aligned at the same time $t^*$, ii) segmented in homogeneous sliding temporal windows $W$, and iii) represented by a feature vector of aggregation functions $\cup = \{\mu, \max, \min\}$. In Fig. 4, we show a visual representation of the segmentation and aggregation of an RSSI sensor stream defined by sliding temporal windows and aggregations of avg, min and max.

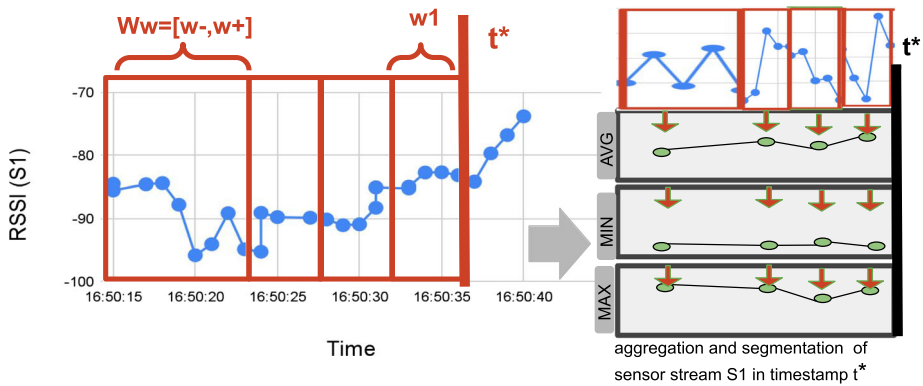

**Fig. 4** Segmentation and aggregation by means, min and max of RSSI sensor stream defined by sliding temporal windows

## 3.3 Deep learning model to estimate inhabitant position from location sensors

We describe the deep learning (DL) model for estimating the position of inhabitants in real time as follows: for each tag $e$ and given timestamp $t$, the input $S_e^*(t^*)$ is configured by a given number of sensor signals $S$, which represent the RSSI collected by the anchor, within a temporal window of steps $W$ configured an input matrix of $S x W$. First, three layers of 1D CNN are integrated as spatial feature extractors, which shape 2, 3, and 3, respectively. Next, two layers of LSTM with 32 units model the temporal dependencies of the upper features of CNN. Dropout layers are included to reduce parameters and to prevent feature co-adaptation and overfitting [73]. The combination of CNN-LSTM Hybrid Networks has been selected because it provides encouraging results in fingerprinting [74]. Third, a multilayer perceptron model of 2 layers of 512 and 256 learns from the spatial and temporal patterns to provide an output, which is the regression of X and Y locations with a sigmoid activation function. Backpropagation learning is applied to the entire network based on the loss metric of the mean squared error and the Adam optimiser. In Fig. 5, we describe the layer configuration in the proposed deep learning model.

In this work, we evaluate the performance of CNN, LSTM and CNN+LSTM (here described). The latter provides the most optimal results from the given DL model architectures. Additionally, for room estimation, the final output of the model has been configured to N outputs that represent the prediction for each room under a classification problem perspective. In this case, the activation function softmax is applied in the output, and the loss metric of binary cross-entropy is selected in back-propagation learning.

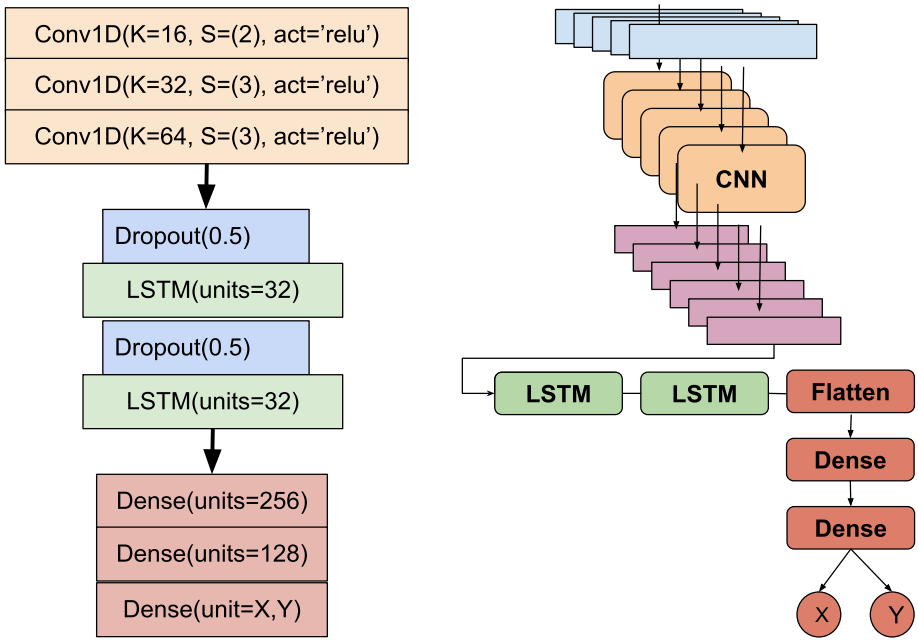

**Fig. 5** Configuration of deep learning models, including CNN and LSTM layers

## 4 Case study

In this section, we describe the case study developed in this work to evaluate and present the use of UWB technology in a real-life context. The devices deployed in this case study are included in the Pozyx Enterprise Kit Lite, described in Section 3. Each anchor sends the RSSI to a given tag in real time via the MQTT –message queuing telemetry transport– protocol, even if the anchors are not visible to each other and do not offer (x, y) positions by trilateration. The data collection of the case study was provided by the user, who wears the UWB tag, who labelled the indoor location in real time by clicking the coordinates using a tablet with a 10-inch screen that displays a map of the flats. In addition, a similar deployment with BLE has been included in both cases to compare the capabilities of BLE and UWB in realistic conditions. The approach used in both case studies has included the installation of sensors evenly to cover all areas of the home and the collection of RSSI information for both UWB (anchors) and BLE (Raspberry Pi configured as a signal receiving beacon). In the case of UWB, the tag has been worn as a necklace. The Amazfit GTS 2e wristband with Bluetooth connectivity has been used for BLE. Still, we note that any activity bracelet with BLE whose MAC (media access control) address is known can be integrated into ongoing work.

### 4.1 Deployment

This case study includes deployment in two flats with different distributions and sizes. In addition, two different device setups were deployed, which are described in the following sections. We focused on providing two different contexts with different deployments and different numbers of anchors for scientific purposes, to evaluate the models and data collection with two different configurations. The approach of evaluating data collection with different locations and sensor configurations under similar conditions is followed in other relevant work [75–77].

#### 4.1.1 Flat A (four anchor setup)

The first case study took place in a 60$m^2$-flat. Four UWB anchors and four BLE beacons covered the main rooms: living room, kitchen, bathroom, and bedroom. Figure 6 left) shows the location of each of the anchors (in green) and BLE beacons (in blue) on floor plan A.

Figure 7 shows the deployment of the UWB and BLE sensors. The map on which the samples were taken was 460 x 753 pixels, while the floor area was 5800 x 9500 mm. We recorded eight tracks of inhabitants doing different routes related to daily human activities within the environment. The tracks ranged from 5 to 10 minutes, described as follows: track 1) 6'14", track 2) 9'32", track 3) 6'55", track 4) 5'0", track 5) 7'16", track 6) 10'50", track 7) 7'5", track 8) 10'34", giving a total of 63 minutes and 23 seconds recorded.

#### 4.1.2 Flat B (six anchors)

The second case study was recorded in a 100$m^2$-flat. Six UWB anchors were placed in the space, distributed as follows: two in the living room, one in the bathroom, one in the kitchen, one in the bedroom, and one in the central corridor. The number of BLE beacons and their location were the same as in the deployment described above. Figure 6 right) shows the location of each of the anchors (in green) and the BLE beacons (in blue) on floor plan B. The map on which the samples were taken was 536 x 621 pixels, while the floor area was 9500 x

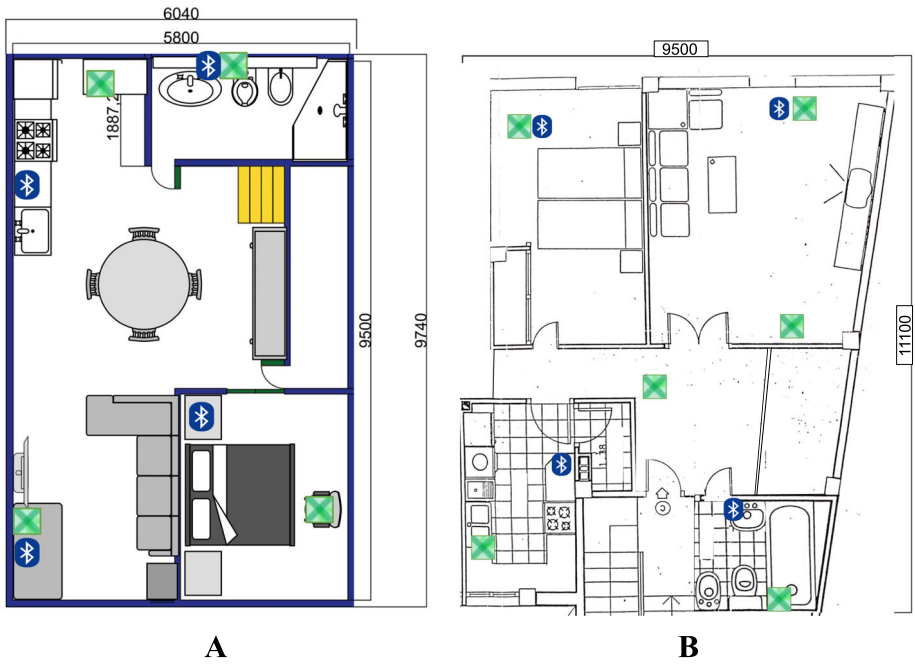

**Fig. 6** left) Floor plan A; right) Floor plan B

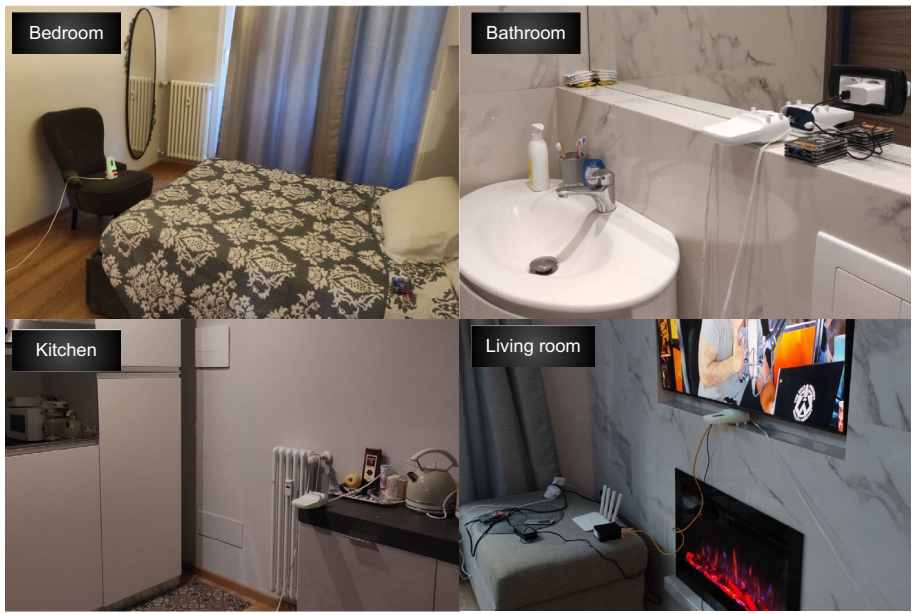

**Fig. 7** Deployment in a real home environment

11100mm. Ten tracks were collected, described as follows: track 1) 3'1", track 2) 2'6", track 3) 2'1", track 4) 1'56", track 5) 2'23", track 6) 2'34", track 7) 2'23", track 8) 1'51", track 9) 2'43", track 10) 1'41" making a total of 22 minutes and 30 seconds of recordings.

### 4.2 Data description

The data used in our model include several key components. First, the location data tagged on the map by the user serves as the ground truth. This data includes the date, the timestamp in milliseconds, and the x and y coordinates in pixels on the map. This precise tagging allows for accurate tracking and validation of the model's predictions. The data received by the UWB anchors are then considered. This data contains the timestamp in milliseconds, the date, the ID of the anchor that sent the signal, and the RSSI value. RSSI values for UWB anchors typically range from -90 to -81. This range indicates the strength of the signal received from the tags, which is crucial to accurately determine the location. Furthermore, data collected from BLE beacons is included. This data set comprises the timestamp in milliseconds, the RSSI value, and the MAC address of the wearable device. The RSSI values for BLE beacons range from -100 to -60, providing another layer of location data through signal strength measurements. Figure 8 shows an example of the data.

For flat A, 16879 RSSI samples were collected and 9176 for flat B. As for the labelled data, 3068 samples were collected for flat A (one sample every second, approx.) and 2658 for flat B (one sample every half second, approx.). In addition, X and Y data location samples were provided by the user who labels the data. A very high frequency of labelling was defined to accurately determine the ground truth ($\mu = 0.98s$, $\sigma = 0.70s$) and ($\mu = 0.41s$, $\sigma = 0.61s$) on flat A and B, respectively. For the synchronisation of RSSI measurements and user-labelled data in real time, the reception and synchronisation of the data are centralised by the MQTT protocol. The RSSI and user labelling data are distributed under publication messages in real time. A subscriber node, whose aim is to synchronise and persist the data, generates a timestamp from the computer clock of the node when the subscription protocol receives messages from RSSI and labelling data. In the learning stage, a time step of one second is defined to compute the points of time that segment the model's RSSI input data (under a sliding window approach). The output location, which is related to each time step and input data, is computed by linear interpolation of the previous and next locations provided by the user. User-tagged data with a difference greater than 5 seconds between samples have been

**Fig. 8** Example of the structure of the collected data. From top to bottom: Labelled data, UWB data and BLE data

```
2022-11-15 15:16:19 1668521779  288 525
2022-11-15 15:16:19 1668521779  288 525
2022-11-15 15:16:24 1668521784  300 520    Labelled
2022-11-15 15:16:26 1668521786  305 527    data
2022-11-15 15:16:26 1668521786  305 527
2022-11-15 15:16:26 1668521786  317 528

1668519249  2022-11-15 14:34:09  4842   -95.53
1668519249  2022-11-15 14:34:09  4908   -90.9
1668519249  2022-11-15 14:34:09  8139   -91.75   UWB
1668519249  2022-11-15 14:34:09  28079  -82.29   data
1668519249  2022-11-15 14:34:09  31176  -84.37
1668519249  2022-11-15 14:34:09  38719  -92.97

1668519251  -74  c8:5b:4f:86:ed:a2
1668519256  -73  c8:5b:4f:86:ed:a2
1668519258  -77  c8:5b:4f:86:ed:a2    BLE
1668519262  -83  c8:5b:4f:86:ed:a2    data
1668519264  -74  c8:5b:4f:86:ed:a2
1668519268  -85  c8:5b:4f:86:ed:a2
```

discarded, treating them as a user break in the labelling stages. 2.43% samples from flat A had these characteristics and were discarded, while only 0.72% samples from flat B were discarded.

### 4.3 Data analysis

This section presents a more visual analysis of the input data we are working with. Figure 9 left) shows how much information has been captured by each anchor (flat B data). It can be seen that the distribution is relatively equal, ranging between 15 and 20% of the total data, which is good because it means that the anchors are all working correctly and we have a sample of data from all parts of the flat. In addition, we can observe the amount of data that have been captured for each room of the flat (Fig. 9 right).

Subsequently, in the next section, the results will show the success rate of the prediction algorithm in each dwelling, which tries to classify the data in a room of the flat and compares it with the real location.

### 4.4 Experimental setup

This section describes the experimental setup, results and conclusions reached in the study. Compared to other approaches, we have included other proposals with related models: i) user room classification based on RSSI signals from UWB and BLE [37], ii) SVM as an RSSI-based fingerprint model for location estimation [42], iii) integration of RF as an RSSI-based fingerprint model for location estimation [43], and iv) integration of DL (CNN + LSTM) as an RSSI-based fingerprint model for location estimation. CNN and LSTM have previously been used separately in [78]. In addition, the CNN+LSTM network enhanced with an attention layer has also been tested.

The results of this hybrid approach (CNN+LSTM) improve fingerprint recognition and matching from collective datasets, as in [79] where they demonstrate good fingerprint recognition with 98% accuracy.

First, we evaluated the DL models (CNN, LSTM, CNN + LSTM, and CNN+LSTM +ATTENTION) to estimate the location of the inhabitants. Second, we analysed the impact on performance using several configurations of time windows with different sizes, including past and current approaches [80]. We have used incremental window sizes [80–82]: i) 1-second subwindows are defined for short-term evaluation of window size (to cover a total window size of 4 and 12 seconds), generating a sequence of 4 and 12 values, respectively; ii)

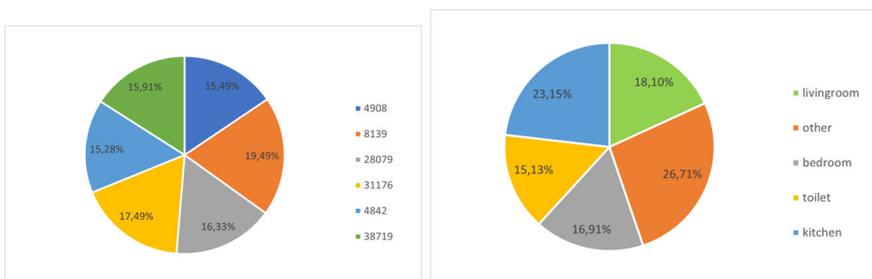

**Fig. 9** left) Percentage data captured by the anchors; right) Data percentage for each room

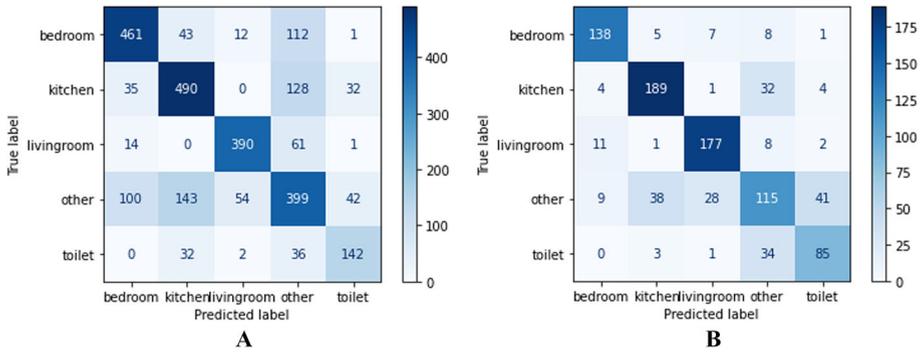

**Fig. 10** Accuracy inhabitant location in rooms from BLE RSSI in flats A left) and B right)

2-second subwindows are defined for long-term evaluation of window size (to cover a total window size of 20 and 30 seconds), generating a sequence of 16 and 21 values respectively.

Third, the mean square error metric and the 10-cross-validation validation method were used to evaluate the location regression. We use the cross-validation method to evaluate the results and ensure that they are independent of the partition between the training and test data, thereby being able to generalise and avoid overfitting. Lastly, CNN + LSTM, which provides the best configuration of DL approaches, was evaluated for the BLE and UWB anchor tag, where a classification problem is computed to estimate the room of the inhabitant [37]. Data samples that did not contain valid RSSI values were discarded because they did not provide information about the location of the user, either because the device was unavailable or because it returned an infinite value, indicating that it was too far away. It is important to note that this has only happened with the RSSI values of BLE; in the case of UWB, RSSI data were always available from the anchors.

In the case of flat A, the algorithm with BLE RSSI data was able to correctly predict the room where the user was with 70% accuracy and 75% accuracy in flat B (see Fig. 10).

In the case of UWB RSSI data, we achieved 85% accuracy for flat A and 83% accuracy for flat B in the room classification, so there is a noticeable improvement (see Fig. 11).

Furthermore, post-prediction, our program computed the total absolute error for variables x and y. We experimented with various deep learning techniques, including CNN, LSTM, a combination of LSTM+CNN, and the same hybrid model configuration enhanced with an attention layer. The computation was performed on data obtained from both flats, yielding identical results. In each case, the LSTM+CNN method proved to be the most effective, followed by the same model with an attention layer (due to the minimal error difference, it has been excluded from the graph), and then by CNN (Fig. 12).

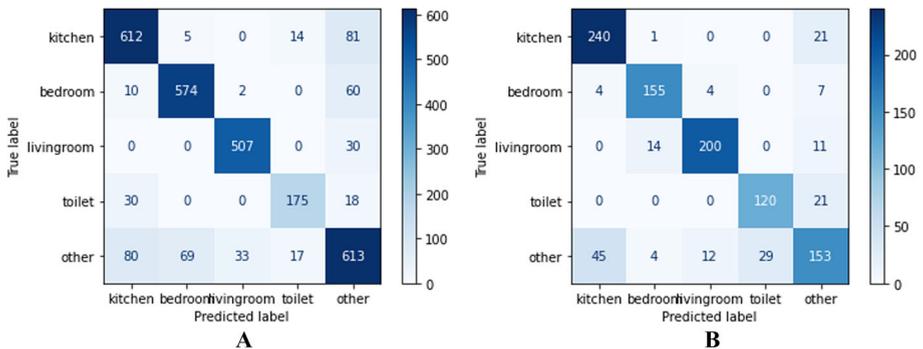

**Fig. 11** Accuracy inhabitant location in rooms from UWB RSSI in flats A left) and B right)

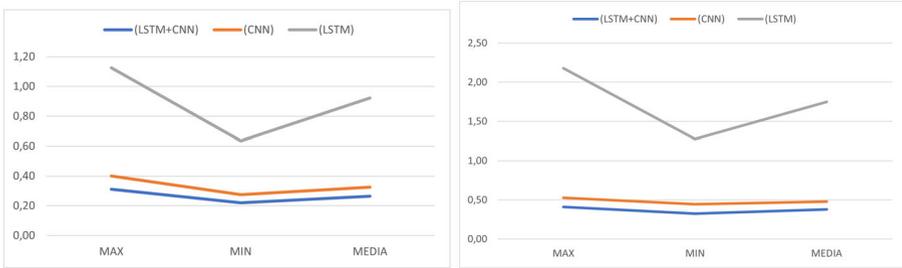

**Fig. 12** Comparison error abs x (left) and abs y (right) with 12s of window

Tables 2 and 3 compare the results obtained, where the column "WINDOWING" indicates whether past and future (PAST+FUTURE) or (ONLY PAST) temporary windows have been used, "MODEL" indicates the model used which in our case can be CNN, LSTM, LSTM+CNN, LSTM+CNN+ATTENTION SVM, RF, and "W. SIZE" shows the window size, and "MAE" contains the value obtained by taking the mean of the absolute error (MAE) of X and Y (in metres). We note that the MAE error is computed between the estimated data of the models and the ground truth provided by the user under high-frequency and real-time conditions. We have also tested changing the size of the time windows and concluded that there is an inversely proportional relationship between window size and error, where increasing the window size reduces the error, but this has diminishing returns: as you increase the window size, the proportional error is reduced less and the execution time is longer (Fig. 13).

The average X-Y error was compared using two configurations: i) past and future time windows, and ii) past window approaches, showing no significant differences when using future windows, as the differences are not more than one point on average, and sometimes only the past window approach is better. The following graphs illustrate this: the blue bar is past + future windows, and the orange bar is only past windows Fig. 14. Table 4 compares the results of the 2D location estimation with CNN + LSTM on the TDOA trilateration obtained from Pozyx, including MAE in the X and Y coordinates and the number of timestamps where the models do not provide the location. This highlights the good performance of the RSSI and DL-based approach and its robustness in estimating location in areas with poor coverage, where trilateration fails. Additionally, a video shows the dynamically displayed location predicted by the system on each flat compared to the one tagged by the user, with red circles for predicted and blue circles for labelled locations. The playback speed of flat A is three times faster than flat B due to the data set duration being 60 versus approximately 20 minutes, respectively. The video is available on YouTube: https://youtu.be/FUluz8Dz3ns.

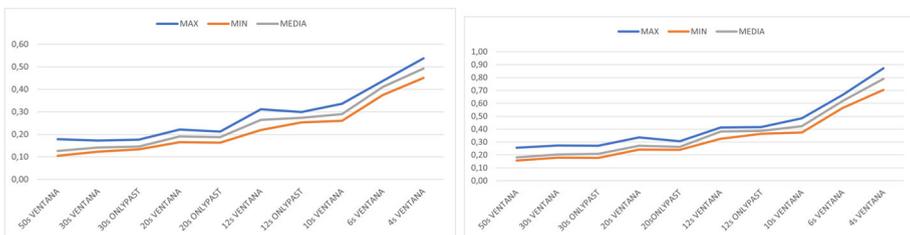

**Fig. 13** Comparison error abs X (left) and abs y (right) LSTM+CNN

**Table 2** Comparison of the results obtained with the different methods on flat A

| WINDOWING | MODEL | W. SIZE | MAE |
|---|---|---|---|
| PAST+FUTURE | CNN | | 0.57 |
| PAST+FUTURE | LSTM | | 1.80 |
| PAST+FUTURE | LSTM+CNN | | 0.64 |
| PAST+FUTURE | LSTM+CNN+ATTENTION | | 0.84 |
| PAST+FUTURE | SVM | 4 | 0.93 |
| PAST+FUTURE | RF | | 0.70 |
| ONLYPAST | SVM | | 0.98 |
| ONLYPAST | RF | | 0.80 |
| PAST+FUTURE | CNN | | 0.40 |
| PAST+FUTURE | LSTM | | 1.33 |
| PAST+FUTURE | LSTM+CNN | | 0.32 |
| PAST+FUTURE | LSTM+CNN+ATTENTION | | 0.41 |
| PAST+FUTURE | SVM | 12 | 0.71 |
| PAST+FUTURE | RF | | 0.55 |
| ONLYPAST | LSTM+CNN | | 0.33 |
| ONLYPAST | LSTM+CNN+ATTENTION | | 0.42 |
| ONLYPAST | SVM | | 0.74 |
| ONLYPAST | RF | | 0.55 |
| PAST+FUTURE | CNN | | 0.33 |
| PAST+FUTURE | LSTM | | 1.40 |
| PAST+FUTURE | LSTM+CNN | | 0.23 |
| PAST+FUTURE | LSTM+CNN+ATTENTION | | 0.29 |
| PAST+FUTURE | SVM | 20 | 0.61 |
| PAST+FUTURE | RF | | 0.53 |
| ONLYPAST | LSTM+CNN | | 0.23 |
| ONLYPAST | LSTM+CNN+ATTENTION | | 0.29 |
| ONLYPAST | SVM | | 0.68 |
| ONLYPAST | RF | | 0.53 |
| PAST+FUTURE | LSTM+CNN | | 0.17 |
| PAST+FUTURE | LSTM+CNN+ATTENTION | | 0.22 |
| PAST+FUTURE | SVM | | 0.58 |
| PAST+FUTURE | RF | | 0.48 |
| ONLYPAST | LSTM+CNN | 30 | 0.18 |
| ONLYPAST | LSTM+CNN+ATTENTION | 30 | 0.23 |
| ONLYPAST | CNN | | 0.28 |
| ONLYPAST | SVM | | 0.64 |
| ONLYPAST | RF | | 0.51 |

Table 3 Comparison of the results obtained with the different methods on floor B

| WINDOWING | MODEL | W. SIZE | MAE |
|---|---|---|---|
| PAST+FUTURE | CNN | | 0.74 |
| PAST+FUTURE | LSTM | | 2.10 |
| PAST+FUTURE | LSTM+CNN | | 0.66 |
| PAST+FUTURE | LSTM+CNN+ATTENTION | | 0.67 |
| PAST+FUTURE | SVM | 4 | 0.72 |
| PAST+FUTURE | RF | | 0.56 |
| ONLYPAST | SVM | | 0.79 |
| ONLYPAST | RF | | 0.53 |
| PAST+FUTURE | CNN | | 0.55 |
| PAST+FUTURE | LSTM | | 1.24 |
| PAST+FUTURE | LSTM+CNN | | 0.37 |
| PAST+FUTURE | LSTM+CNN+ATTENTION | | 0.38 |
| ONLYPAST | LSTM+CNN | 12 | 0.35 |
| ONLYPAST | LSTM+CNN+ATTENTION | | 0.36 |
| PAST+FUTURE | SVM | | 0.79 |
| PAST+FUTURE | RF | | 0.41 |
| ONLYPAST | SVM | | 0.66 |
| ONLYPAST | RF | | 0.47 |
| PAST+FUTURE | CNN | | 0.50 |
| PAST+FUTURE | LSTM | | 0.61 |
| PAST+FUTURE | LSTM+CNN | | 0.32 |
| PAST+FUTURE | LSTM+CNN+ATTENTION | | 0.33 |
| ONLYPAST | LSTM+CNN | 20 | 0.30 |
| ONLYPAST | LSTM+CNN+ATTENTION | | 0.31 |
| PAST+FUTURE | SVM | | 1.08 |
| PAST+FUTURE | RF | | 0.43 |
| ONLYPAST | SVM | | 0.73 |
| ONLYPAST | RF | | 0.49 |
| PAST+FUTURE | LSTM+CNN | | 0.25 |
| PAST+FUTURE | LSTM+CNN+ATTENTION | | 0.26 |
| ONLYPAST | LSTM+CNN | | 0.25 |
| ONLYPAST | LSTM+CNN+ATTENTION | | 0.26 |
| ONLYPAST | CNN | | 0.44 |
| PAST+FUTURE | SVM | 30 | 1.47 |
| PAST+FUTURE | RF | | 0.43 |
| ONLYPAST | SVM | | 0.89 |
| ONLYPAST | RF | | 0.41 |

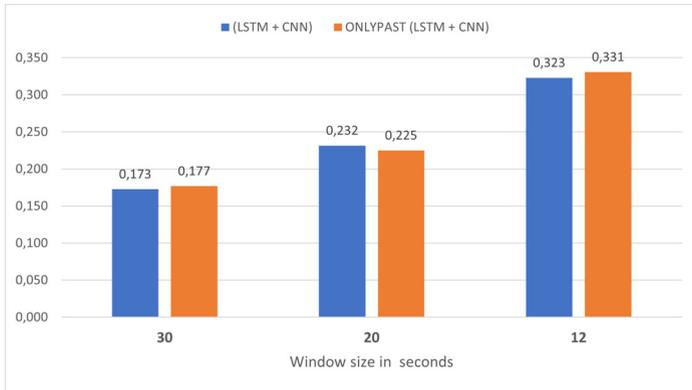

**Fig. 14** Mean error comparison X-Y 30s-20s-12s window

## 5 Conclusion

In this paper, we have evaluated the capabilities of UWB in tracking the daily trajectories of inhabitants in domestic contexts using deep learning models. The necessity of this study arises from the limitations of existing indoor positioning methods, which often face significant accuracy issues due to environmental obstacles and require extensive infrastructure. Traditional positioning methods often struggle in real environments due to obstacles such as walls and furniture, requiring numerous anchors for accurate performance. To address these challenges, we proposed a fingerprint-based approach that calculates location using RSSI values collected from inhabitants in 60 and 100 m$^2$ flats. Our approach offers a more adaptable and cost-effective solution by leveraging deep learning and RSSI-based fingerprinting. This work demonstrates the high accuracy of UWB for indoor positioning, evaluates various deep learning models, and highlights their advantages over traditional methods, suggesting the potential to integrate location data with ambient sensors to improve human activity recognition in multi-occupancy settings and enhance smart home systems. However, it is necessary to test the system in multi-occupancy environments and different architectural layouts to ensure robustness, highlighting areas for future research and development. However, study limitations should be noted. It is necessary to test the system in multi-occupancy environments to determine if signal variability due to the presence of multiple individuals affects the model's performance. In addition, it would be beneficial to evaluate the system in different architectural layouts and building materials to ensure robustness. These limitations highlight areas for future research and development to enhance the applicability and reliability of the proposed solution in diverse real-world scenarios.

**Table 4** Comparison of the UWB results obtained with trilateration+ TDoA vs LSTM+CNN fingerprint + RSSI data

| ALGORITHM | FLAT | MAE (x) | MAE (y) | LOST EST. |
|---|---|---|---|---|
| LSTM+CNN | A | 0.14 | 0.23 | 0 |
| TDoA | A | 0.46 | 0.64 | 4.0% |
| LSTM+CNN | B | 0.20 | 0.30 | 0 |
| TDoA | B | 0.75 | 1.41 | 18.6% |

In the future, our goal is to deploy this architecture in a real environment, operational 24/7, to gather comprehensive data and further refine our models. We will also explore the integration of ambient binary sensors to achieve better HAR in homes with multiple inhabitants. Moreover, we plan to analyse edge computing approaches in which UWB scanning is integrated with wearable devices instead of anchors, potentially reducing infrastructure requirements and enhancing system flexibility. These advancements will contribute to the development of more efficient and user-friendly indoor positioning systems, significantly improving the smart home experience.

**Author Contributions** All authors contributed to writing, reviewing, and approving the final version of the manuscript.

**Funding** Open Access funding provided thanks to the CRUE-CSIC agreement with Springer Nature. This research has been partially funded by the BALLADEER project (PROMETEO/2021/088) from the Consellería de Innovación, Universidades, Ciencia y Sociedad Digital, Generalitat Valenciana and by the AETHER-UA (PID2020-112540RB-C43) project from the Spanish Ministry of Science and Innovation. This work has also been partially funded by "La Conselleria de Innovación, Universidades, Ciencia y Sociedad Digital", under the project "Development of an architecture based on machine learning and data mining techniques for the prediction of indicators in the diagnosis and intervention of autism spectrum disorder. AICO/2020/117". This research has been partially funded by the Spanish Government by the project PID2021-127275OB-I00, FEDER "Una manera de hacer Europa". Lastly, this contribution has been supported by the Spanish Institute of Health ISCIII using the project DTS21-00047.

**Data Availability** The authors declare that the data supporting the findings of this study are available in the article, its supplementary information files, data and the code of the models collected and implemented in this work are available for the research and technical community on GitHub https://github.com/AuroraPR/Indoor-location-tracking.

## Declarations